\documentclass[lettersize,transaction]{IEEEtran}

\normalsize

\IEEEoverridecommandlockouts
\usepackage{cite}
\usepackage{amsmath,amssymb,amsfonts}
\usepackage{algorithmic}
\usepackage{graphicx}
\usepackage{textcomp}
\usepackage{xcolor}
\usepackage{epstopdf} 
\usepackage{epsfig}
\usepackage[super]{nth}
\usepackage{subcaption}
\usepackage{soul}
\usepackage{tabularray}
\usepackage{tabularx}
\usepackage{graphics}
\usepackage{caption}
\usepackage{multirow}
\usepackage{hhline}
\DeclareUnicodeCharacter{2009}{\,}

\def\BibTeX{{\rm B\kern-.05em{\sc i\kern-.025em b}\kern-.08em
    T\kern-.1667em\lower.7ex\hbox{E}\kern-.125emX}}
\begin{document}

\title{Streamlining Multimodal Data Fusion in Wireless Communication and Sensor Networks
}
\author{
    \IEEEauthorblockN{
    Mohammud J. Bocus  \IEEEauthorrefmark{1},
    Xiaoyang Wang       \IEEEauthorrefmark{2},
    Robert. J. Piechocki\IEEEauthorrefmark{1}
}

    \IEEEauthorblockA{\IEEEauthorrefmark{1}Department of Electrical and Electronic Engineering, University of Bristol, UK}\\
    \IEEEauthorblockA{\IEEEauthorrefmark{2}Department of Computer Science, University of Exeter, UK}\\
\{junaid.bocus@bristol.ac.uk, x.wang7@exeter.ac.uk, r.j.piechocki@bristol.ac.uk\}
    }

\maketitle
\begin{abstract}
This paper presents a novel approach for multimodal data fusion based on the Vector-Quantized Variational Autoencoder (VQVAE) architecture. The proposed method is simple yet effective in achieving excellent reconstruction performance on paired MNIST-SVHN data and WiFi spectrogram data. Additionally, the multimodal VQVAE model is extended to the 5G communication scenario, where an end-to-end Channel State Information (CSI) feedback system is implemented to compress data transmitted between the base-station (eNodeB) and User Equipment (UE), without significant loss of performance. The proposed model learns a discriminative compressed feature space for various types of input data (CSI, spectrograms, natural images, etc), making it a suitable solution for applications with limited computational resources.
\end{abstract}


\section{Introduction}
Multimodal fusion is an important aspect of modern artificial intelligence and machine learning systems. It is a process of combining data from multiple sensors to create a comprehensive understanding of the environment. In various applications, such as robotics, autonomous vehicles, and Internet of Things (IoT), multiple sensors are used to capture information from the environment, including vision, audio, lidar, radar, sonar, GPS and more. By combining this data, a more accurate and robust representation of the environment can be created.
Multimodal sensor fusion is important because it helps to overcome the limitations of individual sensors and allows for more reliable and robust decision-making.
However, compression of multimodal data is also needed for increasing efficiency, decreasing the cost of storage and transmission, and facilitating real-time processing of substantial datasets in a variety of applications.

For example, in 5G networks, Channel State Information (CSI) feedback plays a critical role in the communication system. To enhance communication performance, 5G networks make use of sophisticated multi-antenna techniques such as massive MIMO, which necessitate accurate CSI feedback. This feedback is utilized to modify the transmission parameters at the transmitter to account for the fluctuating wireless channel conditions and improve communication quality.
Due to the large number of antennas employed in 5G networks, significant amounts of CSI data are generated. To maintain efficient operation, 5G networks need to apply advanced and smart compression techniques to minimize the size of the CSI feedback data, thereby reducing the latency and overhead of the feedback process.

In the scope of multimodal sensor fusion and compression, 
we propose a multimodal Vector-Quantized Variational Autoencoder (VQVAE) model that can handle multiple modalities within a single model. We first evaluate our straightforward and yet highly efficient model on paired MNIST-SVHN data as a feasibility check for fusion and reconstruction. We then extend our model for two different use cases.
In the first case, we apply the multimodal VQVAE model to WiFi spectrogram data to obtain a compressed and discriminative feature space for passive sensing and Human Activity Recognition (HAR) applications. 
In the second case, the proposed model is evaluated in a 5G communication network perspective, more specifically, we use our model to compress CSI feedback data efficiently that are transmitted from a User Equipment (UE) to a gNodeB (base-station), while maintaining excellent reconstructed channel estimate quality.

This paper is organised as follows: Related works on multimodal data fusion are presented in Section II. 
The background, methodology and system design 
are described in Section III.
Section IV provides detailed information on the experimental setup and corresponding results.
Finally, conclusions are drawn in Section V. 

\begin{figure*}[t]
    \centering
    \includegraphics[width=0.8\textwidth]{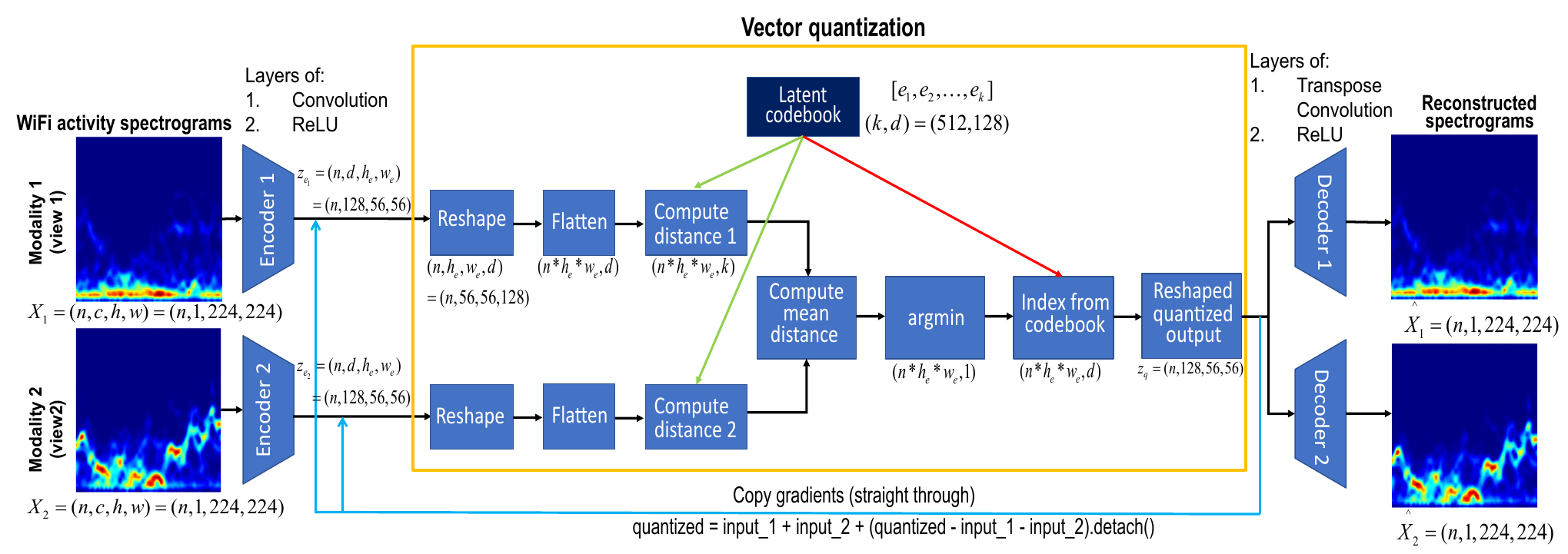}
    \caption{Multimodal VQVAE model.}
    \label{fig:mvqvae}
\end{figure*}

\section{Related Work}
In this section, we review previous research on the topic of multimodal generative modeling and sensor fusion.
The goal of Multimodal Variational Autoencoders (MVAEs) is to learn a joint representation for different kinds of modalities in a self-supervised way, without the need for manual labeling of large amounts of data \cite{latent_fusion}. 
However, obtaining a unified representation from multiple modalities can be challenging as they are often of different data types, having different distributions, levels of sparsity, and dimensions \cite{survey_fusion}. 
To learn a shared representation across multiple modalities, the authors of 
\cite{jmvae} employ a joint inference network. To tackle the challenge of a missing modality, they train individual (single-modal) inference networks for each modality as well as a bi-modal inference network to learn the joint posterior through the use of the Product-of-Experts (PoE) method.
MVAE \cite{mvae}, which is similarly based on PoE, only takes into account a partial combination of the observed modalities, resulting in a smaller number of parameters and increased computing efficiency. On the other hand, the Mixture-of-Experts (MoE) technique is used in \cite{mvae_moe} to learn the shared representation across several modalities.  
To get the best of both worlds, \cite{mopoe} attempts to integrate the benefits of both MoE and PoE in their model, which is referred to as MoPoE (Mixture-of-Products-of-Experts)-VAE.
In \cite{latent_fusion}, the authors proposed a technique for multimodal sensor fusion. The method consists of a two-stage process, whereby a multimodal generative model is trained on unlabelled data in the first stage. Then, in the second stage, this trained generative model serves as a reconstruction prior and the search manifold for various sensor fusion tasks such as multisensory classification, denoising, and recovery from subsampled (compressed) observations.

When it comes to multimodal sensor fusion for human activity detection employing Radio-Frequency (RF), inertial, and/or vision sensors, the bulk of publications have either investigated feature-level fusion or decision-level fusion \cite{latent_fusion}.
For example, \cite{wivi} uses a hybrid Deep Neural Network (DNN) model to perform multimodal fusion at the decision level by exploiting the advantages of both WiFi and vision-based sensors. 
\cite{gimme} describes a method for activity recognition that makes use of four different sensor modalities, including WiFi fingerprints, inertial and motion capture measurements, and skeletal sequences. The measurements from each modality are transformed into images and the fusion of the multimodal data is formulated as a matrix concatenation.
A multimodal Human Activity Recognition (HAR) system that uses WiFi and wearable sensor modalities to jointly infer human behaviours was proposed by the authors in \cite{wiwehar}. They gather measurements of the user's body motions through a wearable Inertial Measurement Unit (IMU) 
and 
WiFi CSI data. Their method consists of calculating the magnitude of the inertial data for each sensor of the IMU and the time-variant Mean Doppler Shift (MDS) from the processed CSI data. The magnitude and the MDS are then independently used to extract different temporal and frequency domain features. The authors adopt the feature-level fusion, whereby feature vectors from the same activity sample are concatenated sequentially. Finally, supervised machine learning methods are employed to classify human activities.
The authors of \cite{armandkassai} leverage the transformer architecture for multimodal sensor fusion. They use different signal processing techniques to extract multiple image-based features from Passive WiFi Radar (PWR) and CSI data such as spectrograms, scalograms and Markov Transition Field (MTF). As compared to the conventional transformer architecture which divides an image into small patches, the authors instead use a different technique whereby each patch represents a different image-based feature.
They developed both supervised and self-supervised models and demonstrated their excellent performance on the HAR task compared to traditional Convolutional Neural Networks (CNNs).
Other approaches used in HAR applications include contrastive learning methods \cite{pretrainpwrcsi,cl_healthcare,cpc},  
which necessitate either multiple views per sensor modality or robust data augmentation methods to generate  pairs of negative and positive samples. 

Recently, some models which leverage discrete representation through vector quantisation have been proposed. Such examples are VQVAE \cite{vqvae} and VQVAE-2 \cite{vqvae2}. 
VQVAE is a type of generative model that combines the principles of autoencoders and vector quantization to generate high-quality, compact representations of data. 
VQVAE is used in various applications, such as image and audio synthesis, to generate high-quality, compressed representations of data that can be used for further analysis or manipulation.
The VQVAE model is made up of three parts; an encoder that converts an image into latent variables, a shared codebook that is used to quantize these continuous latent vectors to a set of discrete latent variables (each vector is replaced with the nearest vector from the codebook), and a decoder that uses the indices of the vectors from the codebook to reconstruct back the image.
VQ-VAE-2 extends the original VQVAE by implementing a two-level hierarchical encoder-decoder model (with multi-scale latent maps) which uses an autoregressive prior, namely, PixelCNN to sample diverse high resolution samples \cite{vqvae2}.

In this work, we extend the VQVAE model to a multimodal setting. More specifically, our proposed model can take multimodal data as input, then it compresses the data to a shared low-dimensional discrete latent representation space and reconstruct the data from the quantized output of the vector quantizer with low reconstruction error using corresponding decoders.

\section{Methodology and System Design}
\subsection{VQVAE}
The VQVAE model consists of an encoder and decoder network, a Vector Quantization (VQ) layer and a reconstruction loss function \cite{vqvae_planning}. The encoder takes as input the data sample $\mathbf{x}$, and outputs the vector $\mathbf{z}_u = f(\mathbf{x})$. The VQ layer maintains an embedding table, $\mathbf{e} \in \mathbb{R}^{k\times d}$, which consists of $k$ vectors of dimension $d$, to quantize the encoder outputs. The VQ layer outputs an index $c$ and the corresponding embedding $\mathbf{e}_c$, which is closest to the input vector  $\mathbf{z}_u$ in Euclidean distance. 
Given an input signal (e.g. an RGB image), the encoder
first encodes it as a $h_e \times w_e \times d$ tensor, where $h_e$ and $w_e$ denote the height and width of the latent representation, respectively. Then, every $d$ dimensional vector is quantized using a nearest-neighbor lookup on the embedding table \cite{online_vq} as per the following: 
\begin{equation}
z_{ij} =  \arg \min_{e \in \mathbf{e}} \parallel \text{enc}(x)_{ij} - e \parallel_2,
\end{equation}
 where $i, j$ refers to a spatial location. 
The decoder then uses the embedding $\mathbf{e}_c$ to reconstruct the input data, $\mathbf{\hat{x}}$. Since the quantization operation is non-differentiable, the gradient is approximated using the straight-through estimator. That is, the gradient from the first layer of the decoder is passed directly to the last layer of the encoder, bypassing the codebook. The latter is updated via exponential moving average of the encoder outputs.
The loss function is represented as
\begin{equation}
\mathcal{L}^t = \mathcal{L}^r(\mathbf{\hat{x}}, \mathbf{x}) + \beta \parallel \mathbf{z}_u - \text{sg} (\mathbf{e}_c) \parallel^2,
\end{equation}
where $\text{sg}(\cdot)$ denotes the stop gradient function. 
The first term is the reconstruction loss while second term is the commitment loss which is used to regularize the encoder to produce vectors close to the embeddings in order to minimize the quantization error \cite{vqvae_planning}. 
The embedding table is updated independently
from the encoder and decoder by minimizing $\min_e \parallel \text{sg} [\text{enc}(x)_{ij}]- e \parallel_2$ \cite{online_vq}.
It should be pointed out that in order to reconstruct the input, only the $h_e \times w_e$ indices are required,  thus achieving a high compression rate.
Compression ratio can be defined as the ratio of the codeword dimension to the original data dimension \cite{9931713}.
For RGB images, the compression rate, $\gamma$, is
given by $\frac{h\times w\times 3\times \log_2(256)}{h_e \times w_e \times \lceil \log_2(k) \rceil}$. 

\begin{figure}[t]
\centering
    \includegraphics[width=0.5\columnwidth]{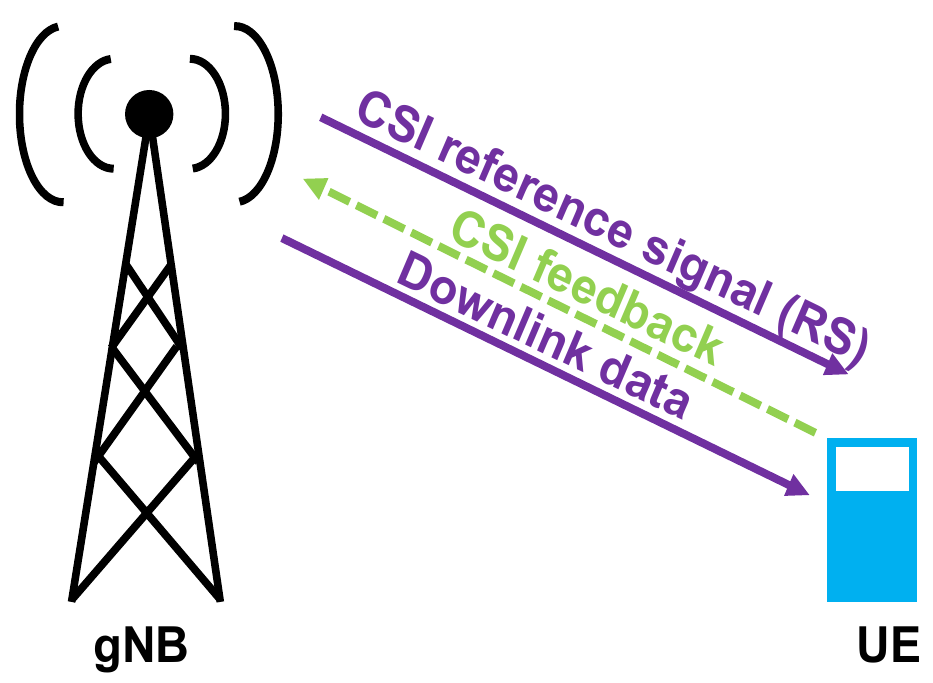}
    \caption{Illustration of communication between a gNodeB (base station) and User Equipment (UE) in a conventional 5G radio network.}
    \label{csi_feedback_illustration}
\end{figure}
\subsection{Proposed Multimodal VQVAE Model}
Our proposed multimodal VQVAE model is shown in Fig. \ref{fig:mvqvae}. Our model follows the same principles as the original VQVAE model. For $M$ input modalities, there will be $M$ encoders and decoders. Each modality data will go through their respective encoders. In the VQ stage, each encoder's output will be reshaped, flattened and the distance is computed using the codebook. Then the mean distance is computed across all input modalities. After the VQ stage, the quantized output serves as input to each corresponding modality decoder to reconstruct their data. We propose a simple but yet very effective framework for multimodal data fusion, as we shall see in section \ref{exp} where we carry out various experiments on paired MNIST-SVHN data and real WiFi spectrogram data, as well as simulated 5G communication CSI feedback data.

\begin{figure*}[t]
\centering
    \includegraphics[width=0.7\linewidth]{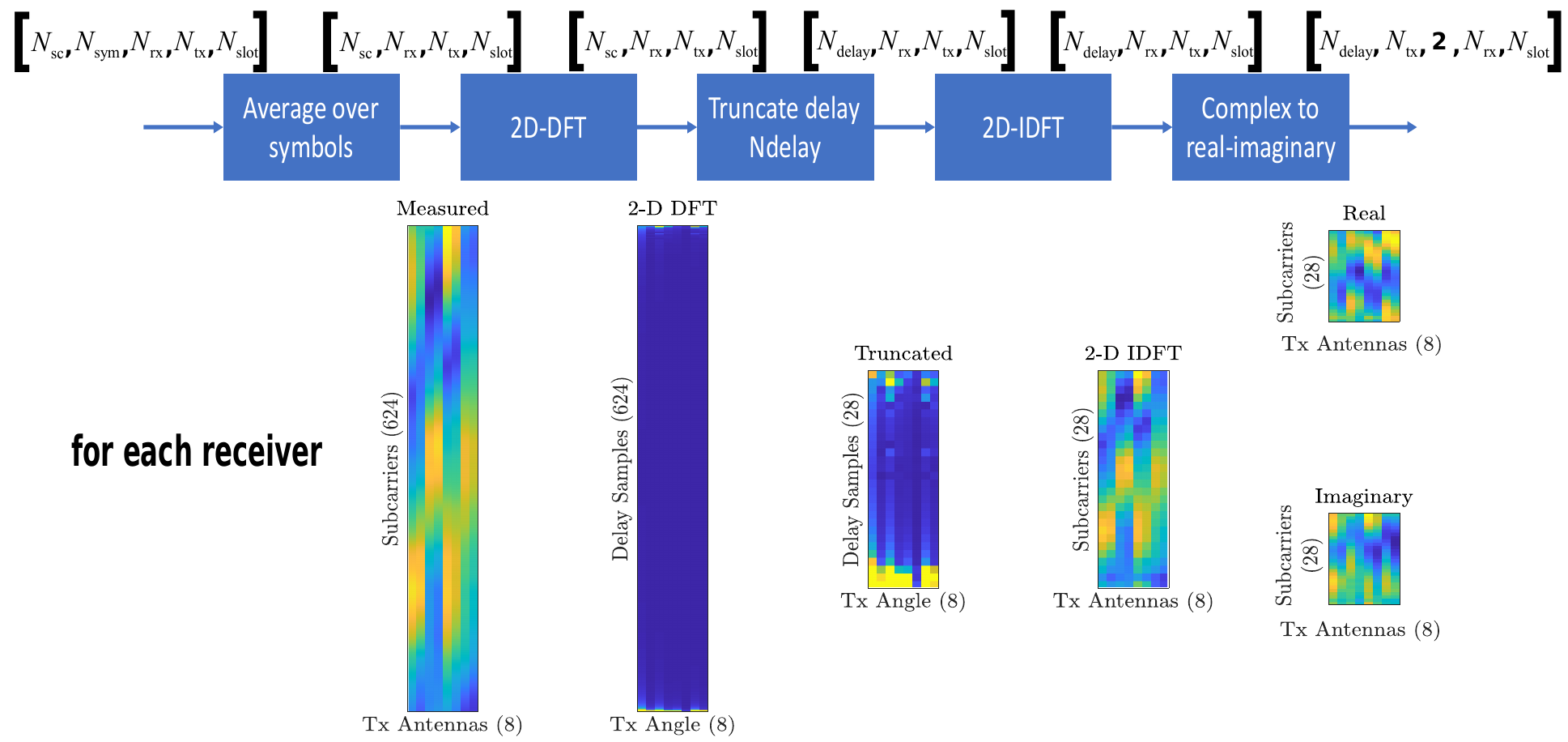}
    \caption{CSI feedback pre-processing steps.}
    \label{csi_feedback_preprocessing}
\end{figure*}

\subsection{WiFi CSI-based Sensing}
\label{wificsi_sec}
WiFi-CSI based sensing systems have been implemented for many applications. For example, human motions within an indoor environment affect the propagation of wireless signals transmitted by the passive WiFi sensors \cite{wificsi}. These applications include HAR \cite{translation_resilient,csipwr,taxonomy}, 
fall detection \cite{falldetection,Damodaran2020}, sign language recognition \cite{languagerecognition}, gesture recognition \cite{8314098,wifinger},    
occupancy detection \cite{FreeDetector}, crowd counting~\cite{ZOU2018309}, respiration monitoring \cite{WiPhone}, among others.
Specific IEEE 802.11 Network Interface Cards (NICs), such as the Intel 5300 \cite{csi_tool} or Atheros \cite{Xie}, can be used to retrieve CSI data.
These WiFi devices leverage Orthogonal Frequency Division Multiplexing (OFDM) at the physical layer. The CSI is represented as a 3D matrix of complex values holding information about the wireless signal characteristics, including propagation delay, amplitude attenuation, and phase shift of multiple propagation paths \cite{armandkassai}. 
WiFi-based sensing is an active area of research. In future real-world large-scale applications, CSI data will be transmitted to a cloud server for computation and data record, thereby creating new challenges for WiFi sensing in terms of reducing communication cost and simultaneously performing model inference \cite{EfficientFi}.

\begin{figure*}[t]
\centering
    \includegraphics[width= \linewidth]{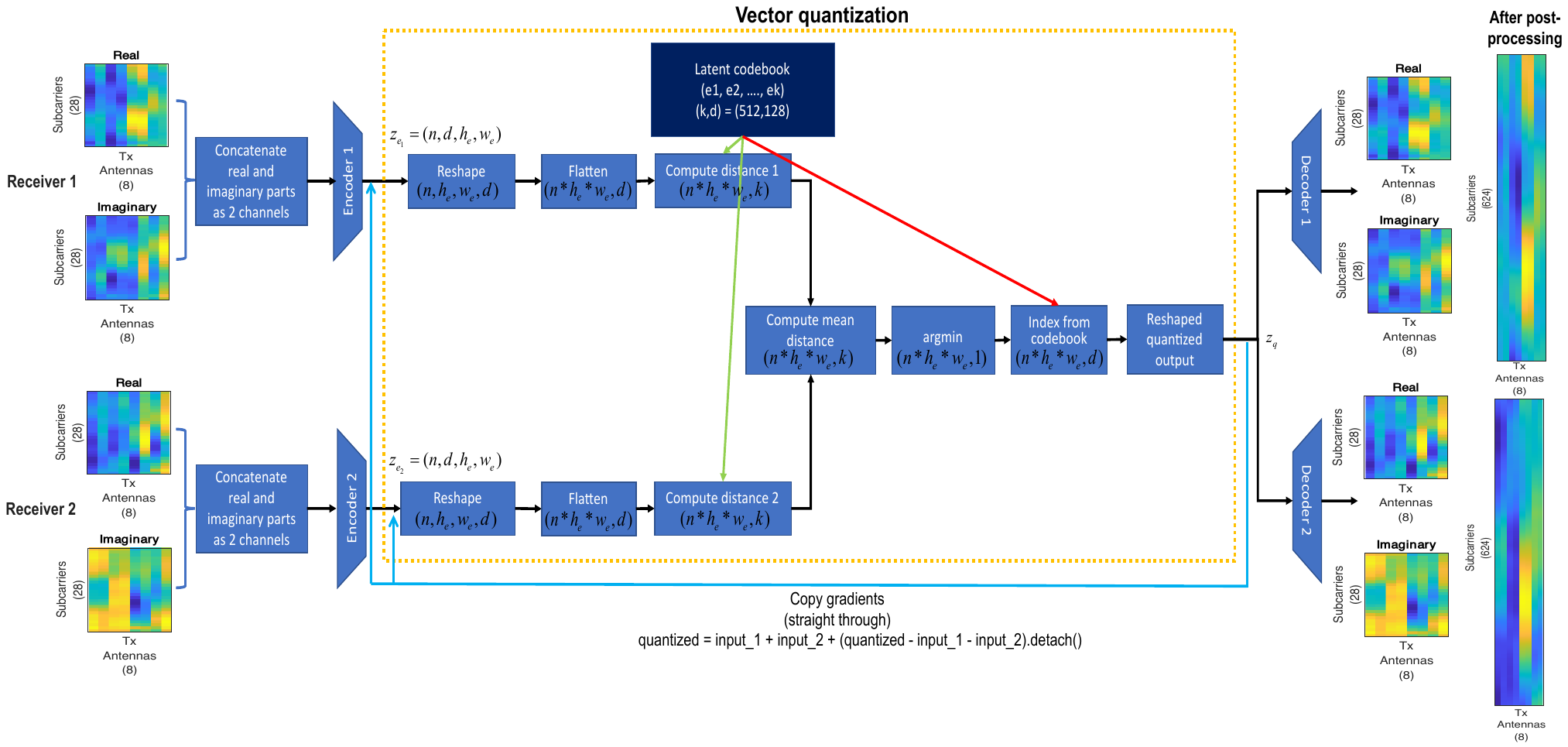}
    \caption{End-to-end CSI feedback multimodal VQVAE model.}
    \label{end_to_end_csi_feedback}
\end{figure*}

The OPERAnet dataset \cite{myoperanetdataset}, which contains freely accessible data from WiFi-based systems, is used in this study. The dataset also contains Kinect and ultra-wideband data.
The dataset was gathered with the goal of analysing HAR and localization methods using data from synchronised RF devices and vision-based sensors.
The various sensors recorded measurements for six human activities carried out by six participants. These activities include sitting down on a chair, 
standing from chair, 
lying down on the floor, 
standing from floor, 
body rotating, 
and walking. 
The aforementioned activities were completed by the participants in two separate rooms at different locations. 
The CSI data were collected across 3 transmit and 3 receive antennas over 30 subcarriers, giving rise to 270 complex CSI values per packet and the sampling rate was set at 1.6 kHz. The data were also captured using two synchronised WiFi CSI receivers.
As a result, a substantial volume of data must be processed. Therefore, the computational complexity of such data may be reduced by using dimensionality reduction techniques like Principal Component Analysis (PCA). Additionally, the resulting data may be subjected to the Short Time Fourier Transform (STFT) to produce spectrograms that resemble those produced by Doppler radars.
The interested reader can learn more about the signal processing pipeline for WiFi CSI in our earlier studies \cite{translation_resilient, csipwr, taxonomy}.
The conversion of raw WiFi CSI data to spectrograms can be regarded as a pre-processing step to data compression. Using the multimodal VQVAE model, the data can be further compressed and used in downstream tasks like human activity classification. Future CSI-based sensing systems will require both a compressed and discriminative feature space for sensing and recognition applications \cite{EfficientFi}.

\subsection {CSI Compression in Communication}
Massive Multiple-Input Multiple-Output (MIMO) technology has been extensively embraced as a top-tier solution for 5G connectivity.
The MIMO system can greatly lessen multi-user interference by utilising CSI at base stations \cite{EfficientFi}.
To do this, the CSI is collected at the UE which is ultimately sent back via a feedback communication link to the base station \cite{6736761}.
The CSI feedback overhead consumes a sizable portion of the uplink bandwidth, especially when there are a lot of transmit antennas.
Many research have been presented to decrease feedback overhead for CSI encoding and decoding in a MIMO system, for example, using  LASSO L1-solver \cite{lasso} or compressive sensing \cite{amp}.
However, because the channel matrix is only roughly sparse, the simple prior cannot completely recover compressed CSI \cite{EfficientFi,csinet}.
In \cite{csinet}, an unsupervised deep learning algorithm (closely related to the autoencoder) is proposed to effectively use the channel structure from training samples in the contexts of CSI sensing and recovery. The algorithm, named CsiNet, basically learns to transform CSI into a near-optimal number of representations (or codewords), and vice-versa (inverse transformation). Comparing CsiNet to current compressed sensing (CS)-based approaches, the reconstruction quality of the recovered CSI is much better.
In \cite{ConvCsiNet}, the authors introduce two new structures, ConvCsiNet (based on a CNN autoencoder) and ShuffleCsiNet (based on ConvCsiNet), for CSI compression. Both structures outperform the previously proposed CsiNet in terms of reconstruction quality as measured by the Normalized Mean Squared Error (NMSE). While ShuffleCsiNet has a lower complexity compared to ConvCsiNet, it still remains more complex than CsiNet.
\begin{figure}[t]
\centering
    \includegraphics[width=0.45\textwidth]{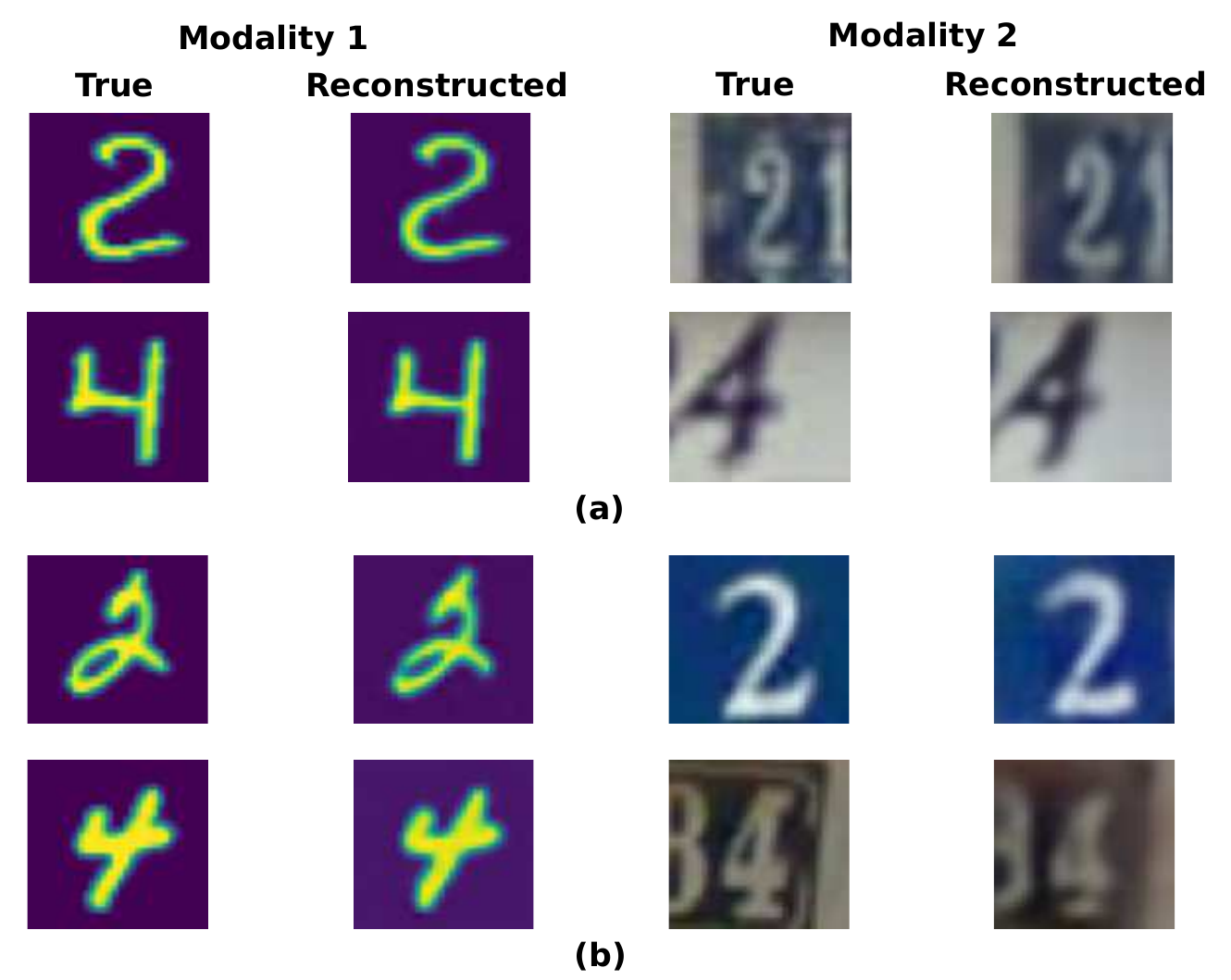}
    \captionsetup{font={normalsize}}
    \caption{Examples of paired MNIST and SVHN images reconstructed using multimodal VQVAE model: (a) $(k,d)=(512,128)$, 
    mean reconstruction error across test data = 0.0033, (b) $(k,d)=(64,128)$,
    mean reconstruction error across test data = 0.0056. }
    \label{fig:mnist_svhn}
\end{figure}
\subsection {CSI Feedback System}
\label{sec:csi_feedback}
In this section, we introduce the concept of CSI feedback for a conventional 5G radio network, as illustrated in Fig. \ref{csi_feedback_illustration}. Our objective is to show how our multimodal VQVAE model in Fig. \ref{fig:mvqvae} can be adapted to compress CSI feedback information (raw channel estimate) over a Clustered Delay Line (CDL) channel.
CSI parameters are values linked to a channel's status that are extracted from the channel estimate array in typical 5G radio networks. These parameters include Rank Indicator (RI), Precoding Matrix Indices (PMI) with different codebook sets and Channel Quality Indication (CQI) \cite{CSIFeedb59:online}. In Fig. \ref{csi_feedback_illustration}, the UE uses the CSI Reference Signal (CSI-RS) to measure and calculate the CSI parameters. The UE then sends (as feedback) the CSI parameters to the base-station (gNodeB) so that the latter can adapt the downlink data transmission in terms of MIMO precoding, number of transmission layers, code rate, and modulation scheme\cite{CSIFeedb59:online}.
In order to reduce the amount of overhead in the CSI feedback data, the UE must process the raw channel estimate. 
While the authors of \cite{csinet,ConvCsiNet} assume a single receiving antenna at the UE, we, on the other hand, propose a model which is applicable to MIMO contexts.
In our approach, we aim for the UE to compress the channel estimate array using a multimodal VQVAE model and then feed it back to the gNodeB. The latter then decompresses and processes the received channel estimate to schedule the downlink data link parameters accordingly.

\subsubsection {5G channel generation}
We use the MATLAB 5G Toolbox$^\text{TM}$ to generate a 5G downlink channel, following the example from \cite{CSIFeedb59:online}. The main parameters used to generate the CDL channel are as follows: RMS Delay spread of 300 ns, maximum Doppler of 5 Hz, 52 resource blocks each consisting of 12 subcarrriers, subcarrier spacing of 15 kHz, 14 symbols per slot, 8 transmit antennas and 2 receive antennas. 
After simulating the channel, the perfect channel estimate matrix, $H_\text{est}$ is represented as an [$N_\text{sc}, N_\text{sym },  N_\text{rx }, N_\text{tx}$] array for each slot, where $N_\text{sc}$, $N_\text{sym }$,  $N_\text{rx }$, and $N_\text{tx}$ correspond to the number of subcarriers, symbols, receive antennas and transmit antennas, respectively.

\begin{table*}
\centering
\caption{Multimodal VQVAE network architecture used in 3 experiments. Conv2d represents 2D convolution and ConvTranspose2d represents 2D transposed convolution. A$\times$(H,W): A denotes the channel number, and
(H,W) represents the height and width of the operation kernel.}
\resizebox{\linewidth}{!}{%
\begin{tblr}{
             rowsep=0pt,
             rows = {font=\linespread{0.8}\selectfont} ,
  width = \linewidth,
  colspec = {Q[115]Q[145]Q[161]Q[131]Q[137]Q[163]Q[181]},
  cells = {c},
  cell{1}{2} = {c=2}{0.236\linewidth},
  cell{1}{4} = {c=2}{0.268\linewidth},
  cell{1}{6} = {c=2}{0.344\linewidth},
  cell{2}{2} = {c=2}{0.236\linewidth},
  cell{2}{4} = {c=2}{0.268\linewidth},
  cell{2}{6} = {c=2}{0.344\linewidth},
  cell{5}{2} = {c=6}{0.848\linewidth},
  cell{9}{2} = {c=2}{0.236\linewidth},
  cell{9}{4} = {c=2}{0.268\linewidth},
  cell{9}{6} = {c=2}{0.344\linewidth},
  hlines,
  vlines,
}
 & \textbf{WiFi CSI (2 receivers)~ } &  & \textbf{Paired MNIST-SVHN } &  & \textbf{CSI feedback (2 receivers) } & \\
{Input to each encoder} & Spectrogram data: (B$\times$1$\times$224$\times$224) &  & {Encoder 1: MNIST: (B$\times$1$\times$32$\times$32)\\Encoder 2: SVHN: (B$\times$3$\times$32$\times$32)  } &  & { Pre-processed CSI feedback data: (B$\times$2$\times$28$\times$8)\\No. of channels = 2 because of real and imaginary components.\\28$\times$8 corresponds to 28 subcarriers and 8 transmit antennas after pre-processing.} & \\
 & \textbf{Encoder} & \textbf{Decoder} & \textbf{Encoder} & \textbf{Decoder} & \textbf{Encoder} & \textbf{Decoder}\\
 & {Conv2d 64$\times$(4,4), stride=(2, 2), padding=(1,1)\\ReLU\\Conv2d 128$\times$(4,4), stride=(2, 2), padding=(1,1)\\ReLU\\Conv2d 128$\times$(3,3), stride=(1, 1), padding=(1,1)~ ~ ~ ~ ~} & Conv2d 128$\times$(3,3), stride=(1, 1), padding=(1,1)~ ~ ~ ~ ~~ & {Conv2d 64$\times$(4,4), stride=(2, 2), padding=(1,1)\\ReLU\\Conv2d 128$\times$(4,4), stride=(2, 2), padding=(1,1)\\ReLU\\Conv2d 128$\times$(3,3), stride=(1, 1), padding=(1,1)~ ~ ~ ~ ~ ~} & {Conv2d 128$\times$(3,3), stride=(1, 1), padding=(1,1)~\\~ ~ ~ ~ ~} & {Conv2d 64$\times$(4,3), stride=(2, 1),
padding=(1,1) 
\\ReLU
\\Conv2d 128$\times$(2,3), stride=(2, 1),
padding=(1,1)~ ~ ~ ~} & Conv2d 128$\times$(3,3), stride=(1, 1),
padding=(1,1)~ ~ ~ ~~\\
{Residual Stack 
\\(no. of residual blocks=2)~~} & {  ReLU
\\Conv2d 32$\times$(3,3), stride=(1, 1),
padding=(1,1) , bias=False
\\ReLU
\\Conv2d 128$\times$(1,1), stride=(1, 1) ,
bias=False
\\
ReLU 
\\Conv2d 32$\times$(3,3), stride=(1, 1),
padding=(1,1) , bias=False
\\ReLU
\\Conv2d 128$\times$(1,1), stride=(1, 1) ,
bias=False~ ~ ~~ } &  &  &  &  & \\
Encoder output dimension & B$\times$128$\times$56$\times$56 &  & B$\times$128$\times$8$\times$8 &  & B$\times$128$\times$8$\times$8 & \\
Pre-VQ-Conv layer & Conv2d 128$\times$(1,1), stride=(1, 1)~ ~ &  & Conv2d 128$\times$(1,1), stride=(1, 1)~ ~ &  & Conv2d 128$\times$(1,1), stride=(1, 1)~ ~ & \\
 &  & {ConvTranspose2d 64$\times$(4,4), stride=(2, 2), padding=(1,1)\\ReLU\\ConvTranspose2d 2$\times$(4,4), stride=(2, 2), padding=(1,1)~ ~} &  & {ConvTranspose2d 64$\times$(4,4), stride=(2, 2), padding=(1,1)\\ReLU\\ConvTranspose2d 2$\times$(4,4), stride=(2, 2), padding=(1,1)~ ~ ~} &  & {ConvTranspose2d 64$\times$(3,3), stride=(2, 1),
padding=(1,1)\\ReLU 
\\ConvTranspose2d 2$\times$(2,3), stride=(2, 1),
padding=(1,1)~ ~ ~~}\\
Compression rate, $\gamma$ & {= 2$\times$(1$\times$224$\times$224$\times$8) / (56$\times$56$\times$9) = 28.44~~\\(consdering $k$=512~embeddings in the codebook, 2 input spectrograms and 8 bits per channel)~} &  & {= [(1$\times$32$\times$32$\times$8) + (3$\times$32$\times$32$\times$8)] / (8$\times$8$\times$9) = 56.89\\(consdering $k$=512~embeddings in the codebook, MNIST data with 1 channel, SVHN data with 3 channels, and 8 bits per channel)~} &  & {=2$\times$(2$\times$28$\times$8$\times$(4$\times$8))~ / (8$\times$8$\times$9) = 49.78\\(considering pre-processed CSI feedback data, 2 receivers each with 2 channels (real and imag),~ $k$=512~embeddings in the codebook, and~one float number occupies 4 bytes=32 bits)} & 
\end{tblr}
}
\label{enc_dec_param}
\end{table*}  

 \subsubsection {CSI feedback pre-processing}
Here, we pre-process the CSI feedback data to reduce its size and then we convert it to a real-valued arrays, as shown in Fig. \ref{csi_feedback_preprocessing}.
In the first  step, we assume that the channel coherence time is much larger than the slot time, and therefore we average the channel estimate over a slot to obtain an [$N_\text{sc}, 1, N_\text{rx },  N_\text{tx }$] array.
In the second step, a 2D Discrete Fourier Transform (DFT) is applied over the subcarriers and transmit (tx) antennas dimensions for each receive (rx) antenna and slot to transform the CSI data into a sparse angular-delay domain \cite{csinet}. 
Since the channel's multipath delay is limited, the delay dimension is truncated to eliminate values that do not hold any information. 
The sampling period on the delay dimension is $T_\text{delay}=1/(N_\text{sc}*F_\text{ss})$, where $F_\text{ss}$ is the subcarrier spacing. The expected RMS delay spread, represented as the number of delay samples, is given by $\tau_\text{RMS}/T_\text{delay}$, where $\tau_\text{RMS}$ is the channel's RMS delay spread in seconds.
Next, the channel estimate is truncated to an even number of samples that is 10 times the expected RMS delay spread.
Using a greater truncation factor value can decrease the performance loss due to pre-processing. However, this increases the number of required training data points, training time and model complexity, and a model with more learnable parameters might not converge to a better solution \cite{CSIFeedb59:online}.
To revert back to the subcarriers-transmit antennas domain (frequency-spatial), a 2D Inverse DFT (IDFT) is applied to the truncated array \cite{9245055}. This process effectively decimates the channel estimate across the subcarrier dimension (from $52\times 12=624$ subcarriers to 28 subcarriers).
Finally, the complex channel estimate is broken down into its real and imaginary parts to obtain an  [$N_\text{delay}, N_\text{tx}, 2$] array for each receiver channel and slot.
In an end-to-end CSI feedback system, the UE utilizes the CSI-RS signal to obtain the channel estimate for one slot, $H_\text{est}$. The pre-processed channel estimate, $H_\text{tr}$
 is obtained using the  steps in Fig. \ref{csi_feedback_preprocessing}, which is then encoded by using the encoder-VQ portion of the multimodal VQVAE model in Fig. \ref{end_to_end_csi_feedback} to generate a compressed array. The latter is decompressed by the decoder portion of the multimodal VQVAE model to obtain $\hat{H}_\text{tr}$, which is then post-processed to produce $\hat{H}_\text{est}$. Post-processing basically consists of the inverse steps in Fig. \ref{csi_feedback_preprocessing} (real-imaginary to complex conversion, 2D-DFT, etc.).

\begin{figure}[t]
\centering
    \includegraphics[width=0.45\textwidth]{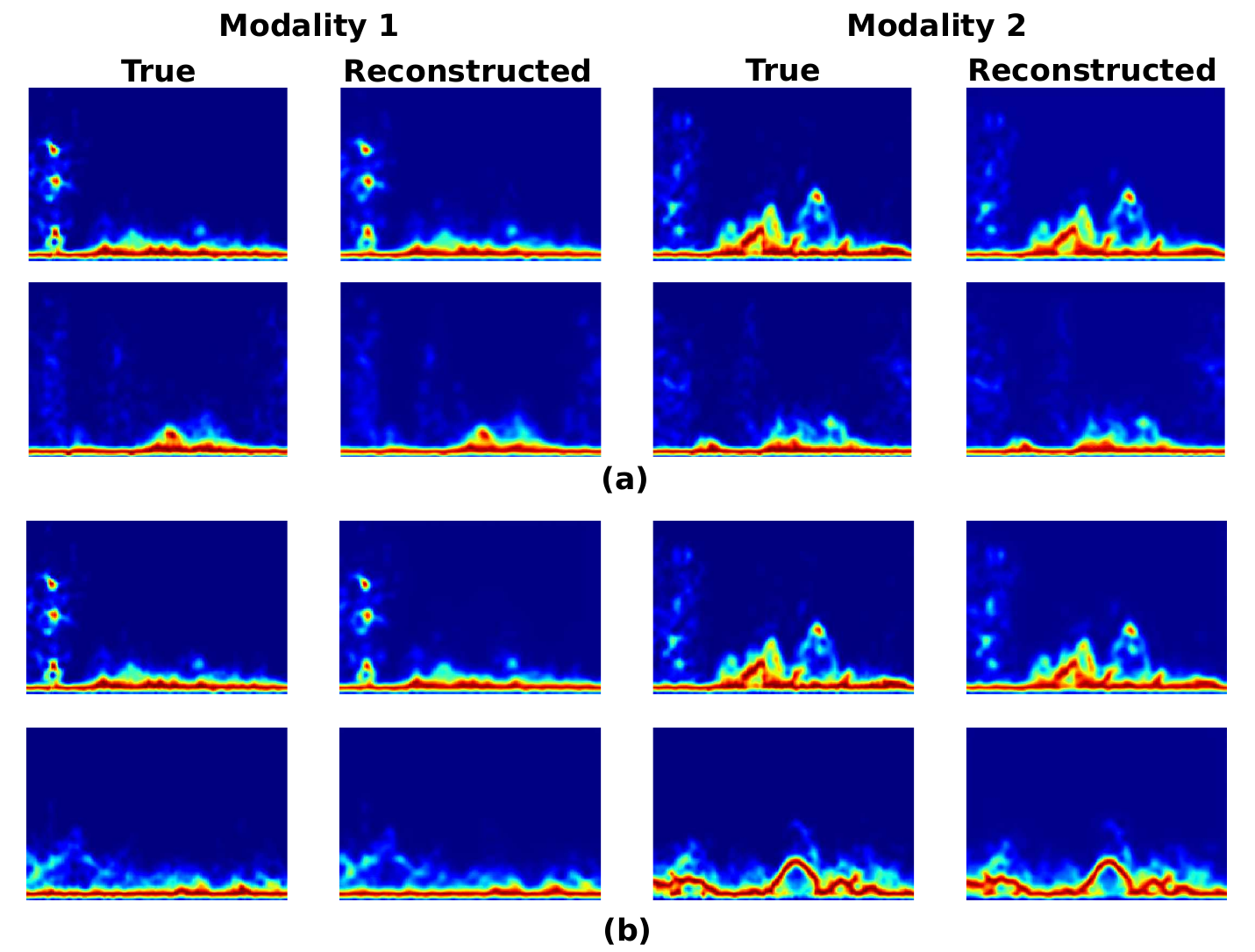}
    \caption{Examples of WiFi human activity spectrograms reconstructed using multimodal VQVAE model: (a)  $(k,d)=(512,128)$,
    mean reconstruction error across test data = 0.0006, (b)  $(k,d)=(64,128)$,
     mean reconstruction error across test data = 0.0006.}
    \label{wifi_spec_ex}
\end{figure}

\begin{figure}[!htp]
\centering
    \includegraphics[width= \columnwidth]{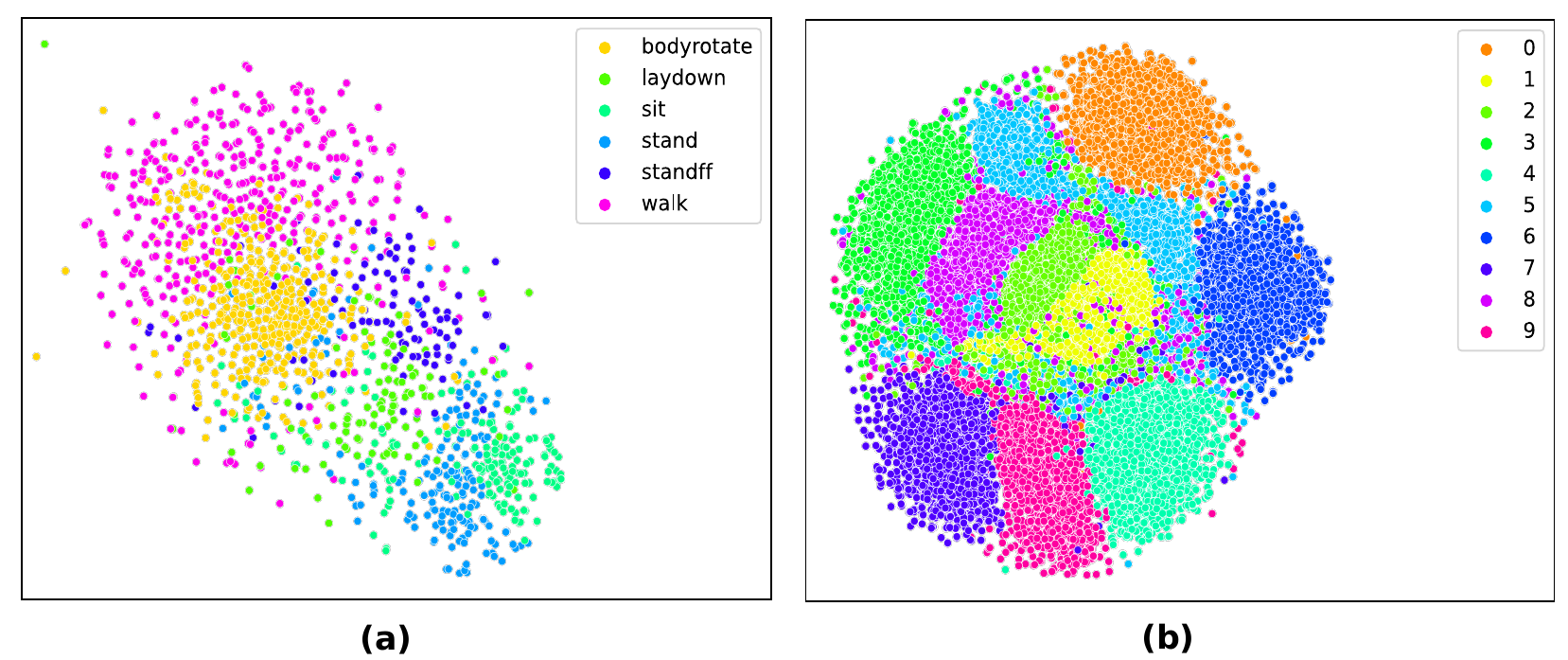}
    \captionsetup{font={normalsize}}
    \caption{t-SNE representation of trained latent space: (a) WiFi spectrogram data, (b) Paired MNIST-SVHN data}
    \label{fig:tsne}
\end{figure}

\begin{figure}[t]
\centering
    \includegraphics[width=\columnwidth]{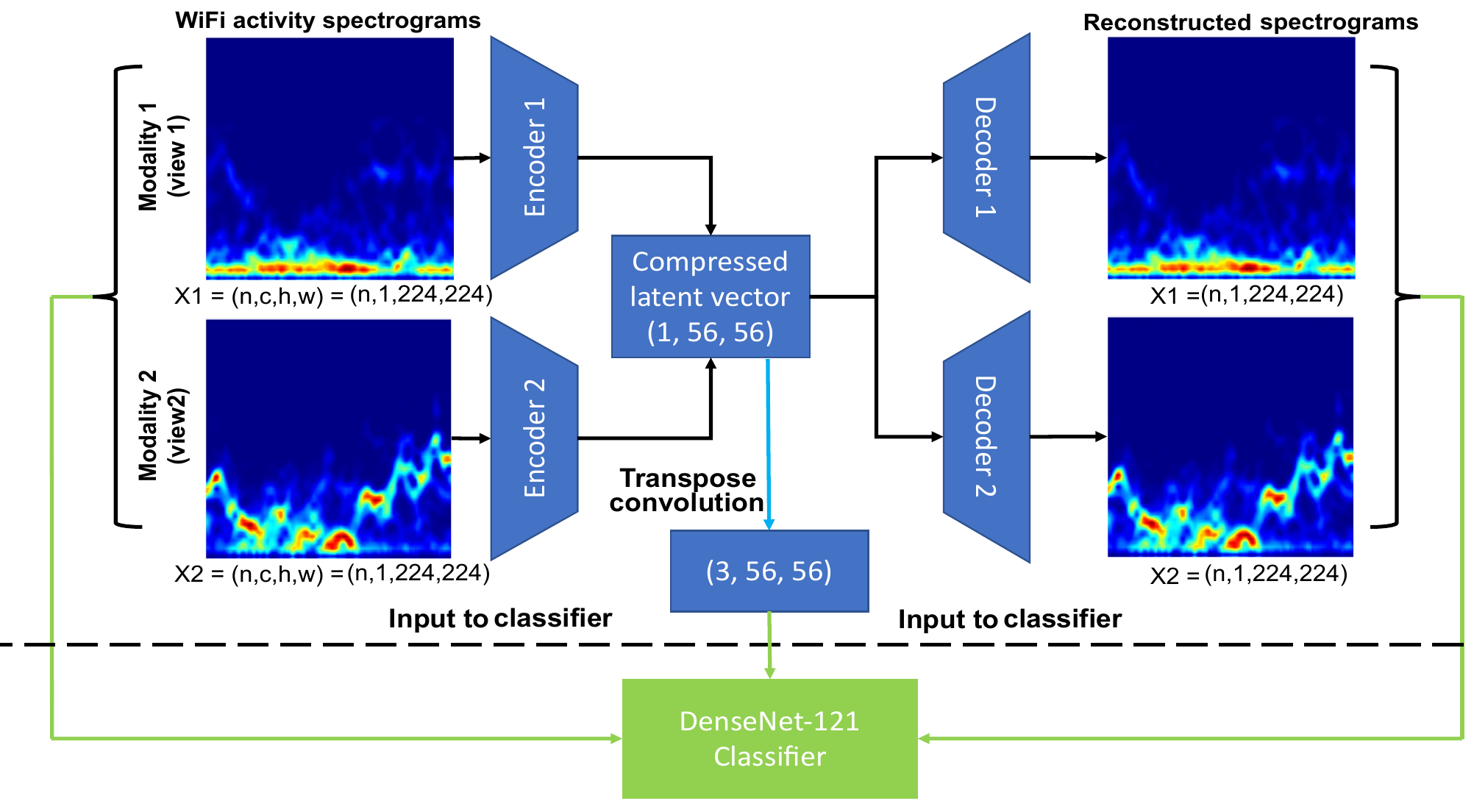}
    \captionsetup{font={normalsize}}
    \caption{WiFi CSI spectrograms classification model.}
    \label{classification_model}
\end{figure}

\begin{table}[t]
\centering
\caption{Classification results using DenseNet-121 on WiFi data.}
\label{tab:my-table}
\begin{tabular}{|c|c|}
\hline
\textbf{Data input type} & \textbf{F1-macro} \\ \hline
\begin{tabular}[c]{@{}c@{}}Original spectrograms (1 channel) \\ Modality 1\end{tabular} & 0.916205 \\ \hline
\begin{tabular}[c]{@{}c@{}}Original spectrograms (1 channel) \\ Modality 2\end{tabular} & 0.936949 \\ \hline
\begin{tabular}[c]{@{}c@{}}Original spectrograms (2 channels) \\ Modality 1 + Modality 2\end{tabular} & 0.947357 \\ \hline
\begin{tabular}[c]{@{}c@{}}Reconstructed spectrograms (2 channels)\\ Modality 1 + Modality 2\end{tabular} & 0.925473   \\ \hline 
Multimodal VQVAE latent vector &  0.905211 \\ \hline 
\end{tabular}%
\end{table}

\begin{figure}[t]
\centering
    \includegraphics[width=\columnwidth]{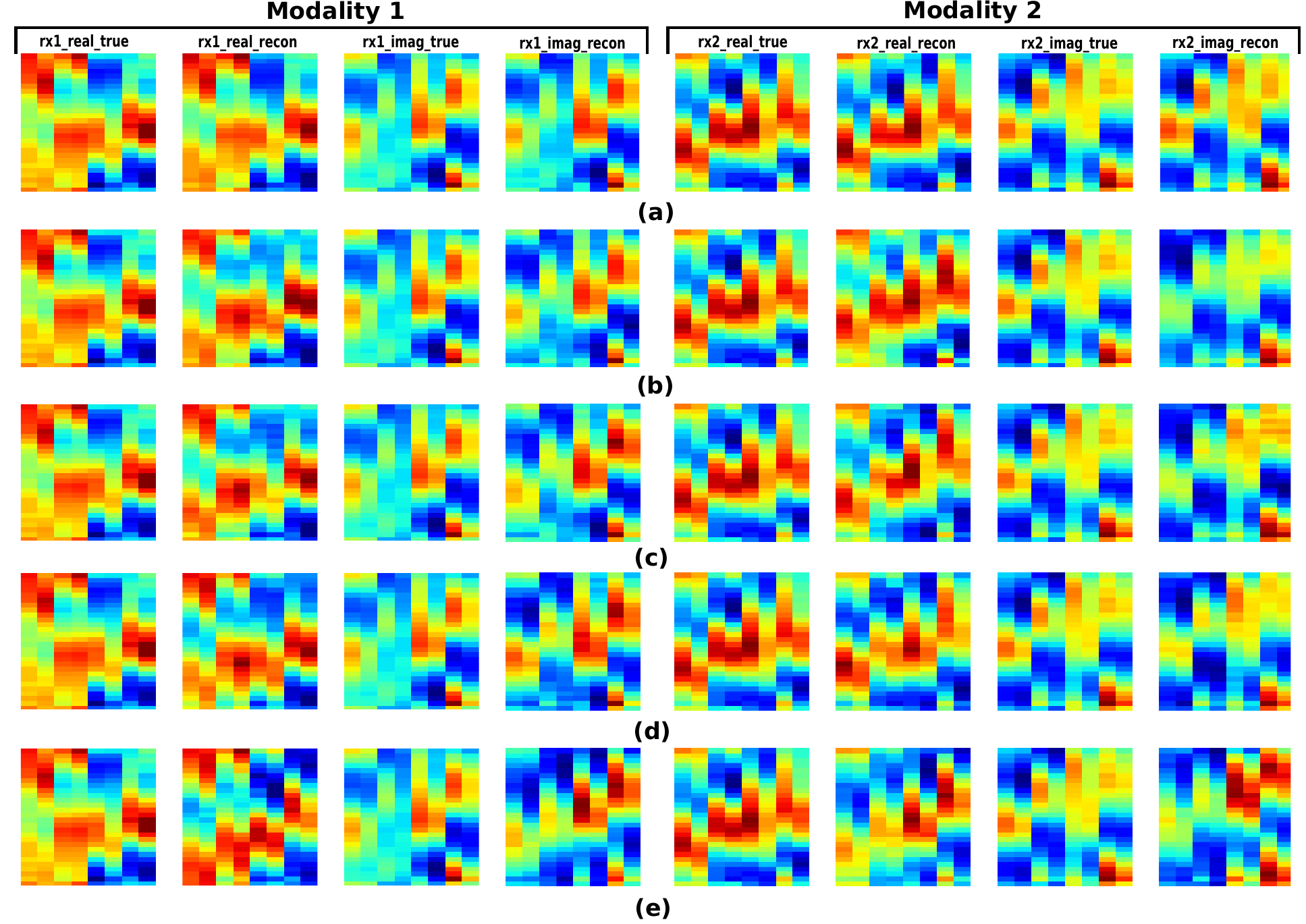}
    \caption{Examples of true and reconstructed CSI feedback samples (real and imaginary) for each receiver (modality) considering different encoder output dimensions: (a) 28$\times$8 (mean reconstruction error across test data = 0.00002062), (b) 8$\times$8  (mean reconstruction error across test data = 0.00006185), (c) 4$\times$8  (mean reconstruction error across test data = 0.00006679), (d) 4$\times$4  (mean reconstruction error across test data = 0.00006738), and (e) 2$\times$2  (mean reconstruction error across test data = 0.00016221).}
    \label{csifeed_ex}
\end{figure}

\begin{figure}[t]
\centering
    \includegraphics[width= \columnwidth]{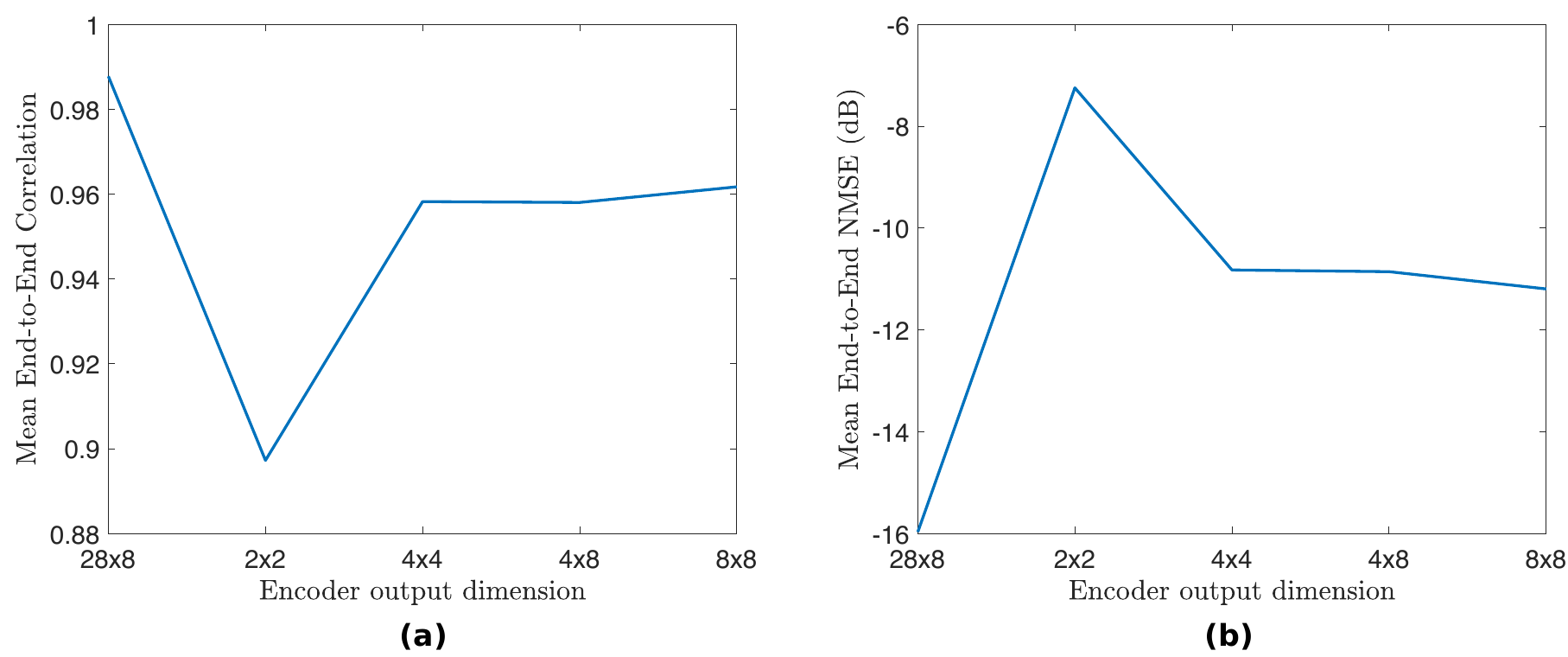}
    \caption{Performance of multimodal VQVAE CSI feedback system considering different encoder output dimensions and $(k,d)=(512,128)$: (a) Mean end-to-end correlation ($\rho$), and (b) Mean NMSE. Encoder output dimension of $28\times8$ refers to uncompressed pre-processed CSI data. Encoder output dimensions of $2\times 2$, $4\times 4$, $4 \times 8$ and $8\times 8$ correspond to compression rates ($\gamma$) of 796.44, 199.11, 99.56 and 49.78, respectively. }
    \label{performance1}
\end{figure}

\begin{figure}[t]
\centering
    \includegraphics[width=\columnwidth]{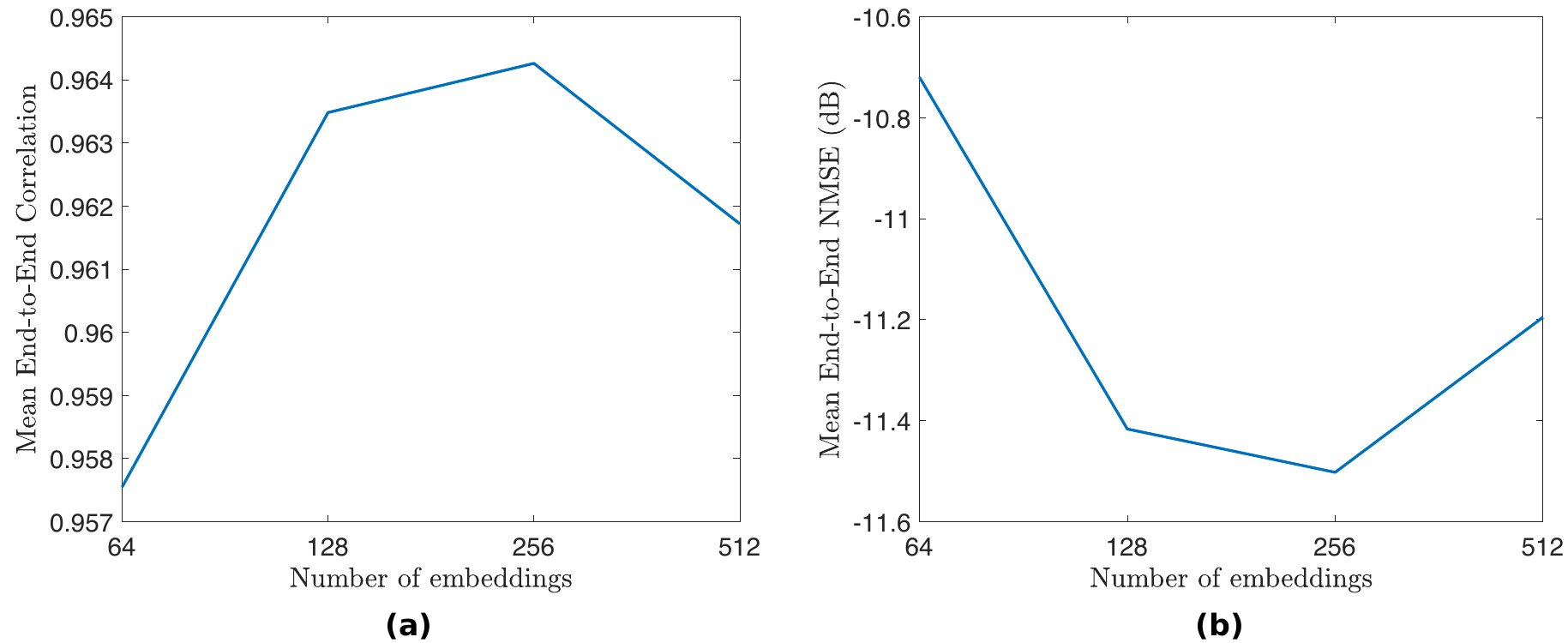}
    \caption{Performance of multimodal VQVAE CSI feedback system for different number of codebook embeddings, considering a fixed embedding dimension ($d$) of 128 and encoder output dimension of $8\times8$ (compression rate, $\gamma$, of 49.78): (a) Mean end-to-end correlation ($\rho$), and (b) Mean NMSE.}
    \label{performance2}
\end{figure}
\begin{figure}[!htp]
\centering
    \includegraphics[width=\columnwidth]{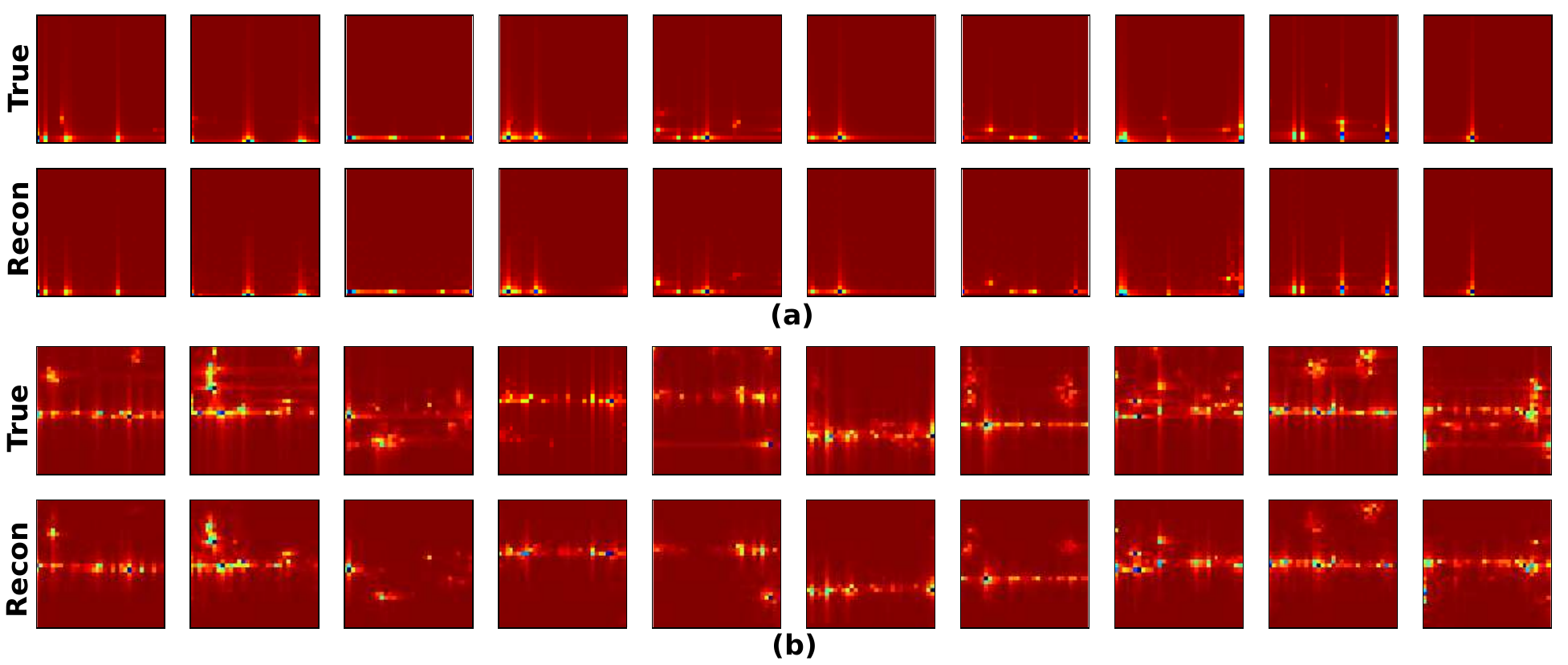}
    \captionsetup{font={normalsize}}
    \caption{Examples of true and reconstructed channel estimate data using multimodal VQVAE model: (a) indoor environment  (mean reconstruction error across test data = 0.00008826), and (b) outdoor environment (mean reconstruction error across test data = 0.00044909). Encoder output dimension is $8\times 8$ and $(k,d)=(256,128)$. 
 }
    \label{csinet_reconstucted_examples}
\end{figure}
\section{Experiments \& Results}
\label{exp}
In this section, the experiments carried out using the multimodal VQVAE architecture are described and the results are presented. The default model's encoder and decoder structures used in each experiment are provided in Table \ref{enc_dec_param}.

\subsection{Paired MNIST-SVHN Data}
We first evaluate the effectiveness of our multimodal VQVAE model on more common datasets, namely, MNIST and SVHN. 
The multimodal dataset is constructed from pairs of MNIST ($1 \times 32 \times 32$) and SVHN ($3 \times 32 \times 32$) images as in \cite{GitHubif20:online}, such that each pair represents the same digit class. 
Each example of a digit class in one dataset is paired randomly with 20 examples of the same digit class from the other dataset.
The training set consists of 50,000 samples while the validation set consists of 48,930 samples. 
The number of embeddings in the codebook is $k=512$ and the embedding dimension is $d=128$. The other parameters used in the model include a commitment cost of 0.25, learning rate of 1e-4, batch size of 64 and 500 epochs for training.
The encoder and decoder structures are given in Table \ref{enc_dec_param}.
The trained latent space for the paired MNIST-SVHN data is shown in Fig. \ref{fig:tsne}(b). 
The trained latent space shows distinct clusters for the 10 digits classes using t-SNE visualization (on the quantized output).
The reconstruction results are shown in Fig. \ref{fig:mnist_svhn}. The mean reconstruction error across the test set is 0.003 when $(k,d)=(512,128)$, and good visual reconstruction quality is observed.
When the number of codebook embeddings is changed from $k=512$ to $k=64$, keeping embedding dimension fixed at $d=128$, that is, the compression rate, $\gamma$, increases from 
from 56.89 to 85.33 considering combined modalities,
the mean reconstruction error also increases.
However, the reconstruction quality is not seriously compromised. 

\subsection{WiFi Spectrogram Data}
\label{wifispec_sec}
In this experiment, we use the WiFi CSI dataset previously described in section \ref{wificsi_sec}. We use various signal processing techniques to convert the raw WiFi CSI data acquired from two synchronised receivers into image-like spectrograms (representing 4s of a human activity), which were resized to a dimension of $1\times 224 \times 224$ (1-channel). As depicted in Fig. \ref{fig:mvqvae}, the spectrogram data from each modality (receiver) serve as input to the encoders in the multimodal VQVAE model. The multimodal data are fused in the VQ stage.
We split the spectrogram data into a training set (60\%=1464 samples), validation set (20\%=488 samples) and test set (20\%=488 samples). The number of embeddings in the codebook is $k=512$ and the embedding dimension is $d=128$. The other parameters used in the model include a commitment cost of 0.25, learning rate of 1e-4, batch size of 64 and 500 epochs for training. Some examples of the reconstruction results using the test set are shown in Fig. \ref{wifi_spec_ex}. We observe very good reconstruction results from both modalities, with a reconstruction error of 0.0006 (across the test set). Moreover, when the codebook size is reduced from 512 to 64 while keeping the embedding dimension fixed at 128, the reconstruction error stays the same. 
That is, even if the compression rate, $\gamma$, increases 
from 28.44 to 42.66 considering combined modalities,
the mean reconstruction error does not increase for the WiFi spectrogram data. This implies that a high level of compression is possible with such data,
without degradation in reconstruction quality.
The trained latent space for WiFi spectrogram data is shown in Fig. \ref{fig:tsne}(a). 
The trained latent space shows distinct clusters using t-SNE visualization (on the quantized output). The model was trained in a self-supervised fashion, and the six clusters in Fig.\ref{fig:tsne}(a) represent the six human activities.

We also evaluate the classification performance of the trained multimodal VQVAE model on the spectrogram data for the purpose of HAR. For this, we use the classification model illustrated in Fig. \ref{classification_model}, 
where training of the DenseNet-121 (pre-trained on ImageNet) is conducted with the original WiFI spectrograms, compressed latent vectors, and reconstructed spectrograms. 
We use an Adam optimizer with
a learning rate of 0.0001. The classifier is trained for 100 epochs.
The classification results, in terms of macro F1-score, are given in Table \ref{tab:my-table}. The classification results on individual modalities are shown in the first and second rows of Table \ref{tab:my-table}. We also combine the original data from both modalities as a 2-channel input data to the classifier. We observe that the macro F1-score for the reconstructed data from the two modalities (treated as a 2-channel input) is slightly lower than with the original input data. 
For the classification on the multimodal VQVAE latent vector, we first apply a 2D transpose convolution on the quantized output 
to obtain a latent vector of dimension $3 \times 56 \times 56$. The resultant latent vector is then used to train the DenseNet-121 classifier. It is observed that the accuracy drops only marginally when classification is performed on the compressed multimodal VQVAE latent vector, which is acceptable because a large compression rate may hinder the discriminative nature of the feature space. 

\subsection{CSI Feedback Data}
Next, we evaluate the multimodal VQVAE model on the CSI feedback data previously described in section \ref{sec:csi_feedback}. 
The model architecture for the CSI feedback data is depicted in Fig. \ref{end_to_end_csi_feedback}, where the CSI data are regarded as a special type of ``image".
7,500 channel realizations were generated, from which 4,500 samples were used for training the multimodal model, and 1,500 samples were used as validation set and 1,500 samples as test set.  
The other parameters used in the model include a commitment cost of 0.25, learning rate of 1e-4, batch size of 64 and 1,000 epochs for training.
The model is trained in an end-to-end manner and the reconstructed data is shown in Fig. \ref{csifeed_ex} for the real and imaginary parts of the CSI data for each modality (receiver). Depending on the encoder output dimension (achieved by modifying the number of layers and kernel size), that is, different compression rate, the mean reconstruction error will vary accordingly. The encoder output dimension of $28\times 8$ is the same as the pre-processed CSI data dimension (refer to Fig. \ref{csi_feedback_preprocessing}), and it is not further compressed by the multimodal VQVAE model. This serves as a baseline and it can be observed from Fig. \ref{csifeed_ex}(a), that this encoder dimension achieves the best mean reconstruction error across the test set. The worst mean reconstruction error is achieved by the encoder with an output dimension of $2\times 2$. After post-processing the real and imaginary components of the reconstructed CSI data from each receiver, we obtain the reconstructed complex channel estimate, $\hat{H}_\text{est}$, with the same dimension as the true channel estimate, $H_\text{est}$, that is $624 \times 2 \times 8$, corresponding to 624 subcarriers, 2 receive antennas and 8 transmit antennas.
The performance of the end-to-end CSI feedback system is shown in Fig. \ref{performance1} for different encoder output dimensions (and thus compression rates) in terms of mean end-to-end correlation, ($\rho$) and Normalized Mean Squared Error (NMSE). 
The cosine correlation, $\rho$, is defined as
\begin{equation}
\rho \triangleq \mathbb{E}\left\{\frac{1}{N_\text{sc}} \sum_{p=1}^{N_\text{sc}} \frac{\left|\hat{\mathbf{h}}_{p}^{H} \mathbf{h}_{p}\right|}{\left\|\hat{\mathbf{h}}_{p}\right\|_2 \left\|\mathbf{h}_{p}\right\|_2}\right\} , 
\end{equation}
where $\mathbf{h}_{p}$ and $\hat{\mathbf{h}}_{p}$ are the true and reconstructed channel estimates of the $p$th subcarrier, respectively. NMSE is computed as
\begin{equation}
\mathrm{NMSE} \triangleq \mathbb{E}\left\{\frac{\|\mathbf{H}-\hat{\mathbf{H}}\|_{2}^{2}}{\|\mathbf{H}\|_{2}^{2}}\right\},
\end{equation}
where $\mathbf{H}$ and $\hat{\mathbf{H}}$ denote the true channel and recovered channel, respectively.
The number of embeddings ($k$) in the codebook and embedding dimension ($d$) are 512 and 128, respectively. It can be observed from Fig. \ref{performance1} that as the compression rate increases, the performance also degrades as expected. However, considering the encoder output dimensions of $4\times 4$, $4 \times 8$ and $8\times 8$, which correspond to compression rates of 199.11, 99.56 and 49.78, respectively, we observe that there is only very slight difference in their performance.
Furthermore, in Fig. \ref{performance2} we analyse the performance of the end-to-end CSI feedback system in terms of different number of embeddings, $k$, in the codebook. An encoder with output dimension of $8\times 8$ (compression rate of 49.78) is considered and the embedding dimension ($d$) is 128. It can be observed that when $k=256$, the best performance is achieved in terms of end-to-end correlation ($\rho$) and NMSE. This means that we can use a smaller number of embeddings ($k$) in the codebook to achieve an even higher compression rate ($\gamma$) with still a better performance.





\subsection{Comparison with State-of-the-Art CSI Feedback Models}
In this experiment, we compare our multimodal VQVAE model to CsiNet\cite{csinet} and ConvCsiNet/ShuffleCsiNet \cite{ConvCsiNet} using the same wireless channel data \cite{GitHubsy82:online}.
The training and testing samples are generated for indoor (picocellular scenario at the 5.3 GHz band) and outdoor (rural scenario at the 300 MHz band) environments using the COST 2100 channel model \cite{COST}.
The base-station is located at the centre of a square of dimension $20$ m$\times$$20$ m and $400$ m$\times$$400$ m for the indoor and outdoor environments, respectively. The UE in each environment is randomly positioned within the measurement area and is equipped with a single receiving antenna. The base-station consists of a Uniform Linear Array (ULA) of $N_\text{tx}=32$ antennas and the number of subcarriers is $N_\text{sc}=1024$. When the channel matrix is converted to the angular-delay domain using 2D-DFT, only the first 32 rows of the channel matrix, $\mathbf{H}$, are retained, thus resulting in a dimension of $32\times 32$ \cite{csinet}.
The training, validation, and testing sets consist of 100,000, 30,000, and 20,000 samples, respectively. We trained our multimodal VQVAE model in Fig. \ref{end_to_end_csi_feedback} for 1,000 epochs, with a learning rate of 1e-4 and batch size of 64. Note that since there is only one receiving antenna in this case, the real and imaginary parts of the channel matrix, $\mathbf{H}$, were considered as two modalities. Therefore, each encoder in Fig. \ref{end_to_end_csi_feedback} takes as input a $1\times 32 \times 32$ channel data (real and imaginary separately). We use the same encoder/decoder structure as in Table \ref{enc_dec_param} for the WiFi CSI data. Each encoder output thus has a dimension of $8\times 8$. We consider an embedding dimension $d=128$ and number of embeddings, $k=256$ and $k=512$ in the evaluation of the correlation, $\rho$, and NMSE in Table \ref{tab:csinet_vs_ours}. 
Note that the channel matrices (true and reconstructed) are transformed back to their original dimensions when computing these two metrics. The compression rates, $\gamma$, are 114 and 128 for $k=512$ and $k=256$, respectively. 
We also include $\gamma=32$ for an encoder output dimension of $16\times 16$ and $(k,d)=(256,128)$. 
The results in Table \ref{tab:csinet_vs_ours} show that the overall performance of our model is much better than CsiNet, ConvCsiNet and ShuffleCsiNet.
For example, considering the same compression rate of $\gamma=32$, our model achieves higher correlation values ($\rho$) and better NMSE than the other models
for both the indoor and outdoor environments.
Also, when we use a compression rate as high as $\gamma=128$, our model still performs better than the other models which use a four times lower compression rate of $\gamma=32$.
Some samples of the true and reconstructed channel data are shown in Fig. \ref{csinet_reconstucted_examples} for the indoor and outdoor environments where good reconstruction performance is observed across the test set samples.
For instance, when the encoder output dimension is $8\times 8$ and $(k,d)=(256,128)$, the mean reconstruction errors across the test data are 0.00008826 and 0.00044909 for the indoor and outdoor environments, respectively. 

\begin{table}
\centering
\caption{Performance comparison between CsiNet, ConvCsiNet, ShuffleCsiNet and our multimodal VQVAE model on the same wireless channel data.}
\label{tab:csinet_vs_ours}
\resizebox{\linewidth}{!}{%
\begin{tblr}{
  cells = {c},
  cell{1}{1} = {r=2}{},
  cell{1}{2} = {r=2}{},
  cell{1}{3} = {c=2}{},
  cell{1}{5} = {c=2}{},
  cell{3}{1} = {r=4}{},
  cell{7}{1} = {r=2}{},
  cell{9}{1} = {r=2}{},
  cell{11}{1} = {r=3}{},
  vlines,
  hline{1,3,7,9,11,14} = {-}{},
  hline{2} = {3-6}{},
  hline{4-6,8,10,12-13} = {2-6}{},
}
\textbf{Method } & \textbf{Compression rate, $\gamma $} & \textbf{Cosine similarity, $\rho$} &  & \textbf{NMSE} & \\
 &  & \textbf{\textbf{Indoor}} & \textbf{\textbf{Outdoor}} & \textbf{\textbf{Indoor}} & \textbf{\textbf{\textbf{\textbf{Outdoor}}}}\\
CsiNet \cite{csinet} & 4 & 0.99 & 0.91 & -17.36 & -8.75\\
 & 16 & 0.93 & 0.79 & -8.65 & -4.51\\
 & 32 & 0.89 & 0.67 & -6.24 & -2.81\\
 & 64 & 0.87 & 0.59 & -5.84 & -1.93\\
ConvCsiNet \cite{ConvCsiNet} & 16 & 0.98 & 0.85 & -13.79 & -6.00\\
 & 32 & 0.95 & 0.82 & -10.10 & -5.21\\
ShuffleCsiNet \cite{ConvCsiNet} & 16 & 0.97 & 0.82 & -12.14 & -5.00\\
 & 32 & 0.94 & 0.74 & -9.41 & -3.50\\
Ours & 32 & 0.98 & 0.93 & -14.52 & -9.99\\
 & 114 & 0.96 & 0.85 & -10.63 & -6.06\\
 & 128 & 0.95 & 0.84 & -10.41 & -5.56
\end{tblr}
}
\end{table}


\section{Conclusion}
In conclusion, our proposed multimodal VQVAE architecture is a highly efficient solution for multimodal data fusion and compression. Its simple structure and excellent performance on paired MNIST-SVHN data, WiFi spectrogram data, and CSI feedback data in a massive MIMO system demonstrate its potential for a wide range of applications. The added compression achieved by the model without severe degradation in reconstruction performance is particularly beneficial in bandwidth-limited scenarios.


\section*{Conflict of interest statement} 
The authors declare that they have no conflict of interest.

\section*{Data availability statement}
The WiFi CSI  spectrogram data have been generated using the dataset which is openly available in figshare at https://doi.org/10.6084/m9.figshare.c.5551209.v1 \cite{myoperanetdataset}. We use the torchvision package to import commonly used datasets such as MNIST and SVHN. The simulated CSI feedback data were generated using the MATLAB 5G Toolbox. The data used in the CsiNet or ConvCsiNet/ShuffleCsiNet systems can be accessed from \cite{GitHubsy82:online}.

\section*{Funding information}
This work was performed as a part of the OPERA Project, funded by the UK Engineering and Physical Sciences Research Council (EPSRC), Grant EP/R018677/1. This work has also been funded in part by the Next-Generation Converged Digital Infrastructure (NG-CDI) Project, supported by BT and Engineering and Physical Sciences Research Council (EPSRC), Grant ref. EP/R004935/1.

\bibliographystyle{IEEEtran}
\bibliography{bibliography}
\end{document}